\documentclass[10pt,twocolumn,letterpaper]{article}

\usepackage[pagenumbers]{cvpr} % To force page numbers, e.g. for an arXiv version

\usepackage{graphicx}
\usepackage{amsmath}
\usepackage{amssymb}
\usepackage{booktabs}
\usepackage{xspace}
\usepackage{interval}
\usepackage{bm}
\usepackage{mathtools}
\usepackage{footmisc}

\usepackage{siunitx}
\usepackage{bbold}
\usepackage{makecell}
\usepackage[dvipsnames]{xcolor}
\definecolor{citecolor}{RGB}{0, 113, 188}
\usepackage[pagebackref=true,breaklinks=true,letterpaper=true,colorlinks,citecolor=citecolor, bookmarks=false]{hyperref}

\newcommand{\veryshortarrow}[1][3pt]{\mathrel{%
   \hbox{\rule[\dimexpr\fontdimen22\textfont2-.2pt\relax]{#1}{.4pt}}%
   \mkern-4mu\hbox{\usefont{U}{lasy}{m}{n}\symbol{41}}}}
 
\usepackage[capitalize]{cleveref}
\crefname{section}{Sec.}{Secs.}
\Crefname{section}{Section}{Sections}
\Crefname{table}{Table}{Tables}
\crefname{table}{Tab.}{Tabs.}

\def\cvprPaperID{****} % *** Enter the CVPR Paper ID here
\def\confName{CVPR}
\def\confYear{2022}

\begin{document}

\title{AdaViT: Adaptive Vision Transformers for Efficient Image Recognition}

\author{
  Lingchen Meng$^{1}$\footnotemark[1]~\hspace{10pt}
  Hengduo Li$^{2}$\footnotemark[1]~\hspace{10pt}
  Bor-Chun Chen$^{3}$\hspace{10pt}
  Shiyi Lan$^{2}$\hspace{10pt} \\
  Zuxuan Wu$^{1}$\footnotemark[2]\hspace{10pt}
  Yu-Gang Jiang$^{1}$\footnotemark[2]~\hspace{10pt}
  Ser-Nam Lim$^{3}$\hspace{10pt}
  \\
$^{1}$Fudan University
\qquad $^{2}$University of Maryland
\qquad $^{3}$Meta AI 
}
\maketitle

\renewcommand{\thefootnote}{\fnsymbol{footnote}}
\footnotetext[1]{Equal contributions.}
\footnotetext[2]{Corresponding authors.}

%%%%%%%%% ABSTRACT
\begin{abstract}
Built on top of self-attention mechanisms, vision transformers have demonstrated remarkable performance on a variety of vision tasks recently. While achieving excellent performance, they still require relatively intensive computational cost that scales up drastically as the numbers of patches, self-attention heads and transformer blocks increase. In this paper, we argue that due to the large variations among images, their
need for modeling long-range dependencies between patches differ. To this end, we introduce AdaViT, an adaptive computation framework that learns to derive usage policies on which patches, self-attention heads and transformer blocks to use throughout the backbone on a per-input basis, aiming to improve inference efficiency of vision transformers with a minimal drop of accuracy for image recognition. Optimized jointly with a transformer backbone in an end-to-end manner, a light-weight decision network is attached to the backbone to produce decisions on-the-fly. Extensive experiments on ImageNet demonstrate that our method obtains more than $2\times$ improvement on efficiency compared to state-of-the-art vision transformers with only $0.8\%$ drop of accuracy, achieving good efficiency/accuracy trade-offs conditioned on different computational budgets. We further conduct quantitative and qualitative analysis on learned usage polices and provide more insights on the redundancy in vision transformers.
\end{abstract}

%%%%%%%%% BODY TEXT

%%%%%%%%% Introduction
\section{Introduction}

Transformers~\cite{transformer}, the dominant architectures for a variety of natural language processing (NLP) tasks, have been attracting an ever-increasing research interest in the computer vision community since the success of the Vision Transformer (ViT)~\cite{vit}. Built on top of self-attention mechanisms, transformers are capable of capturing long-range dependencies among pixels/patches from input images effectively, which is arguably one of the main reasons that they outperform standard CNNs in vision tasks spanning from image classification~\cite{touvron2021training_deit,t2t,chen2021crossvit,li2021localvit,wu2021cvt,liu2021swin,han2021transformer_tnt} to object detection~\cite{liu2021swin,wang2021pyramid,carion2020end_detr,chu2021twins}, action recognition~\cite{fan2021multiscale,liu2021videoswin,zhang2021vidtr} and so forth.

\begin{figure}[!t] \centering
    \resizebox{\linewidth}{!}{\includegraphics[width=\linewidth]{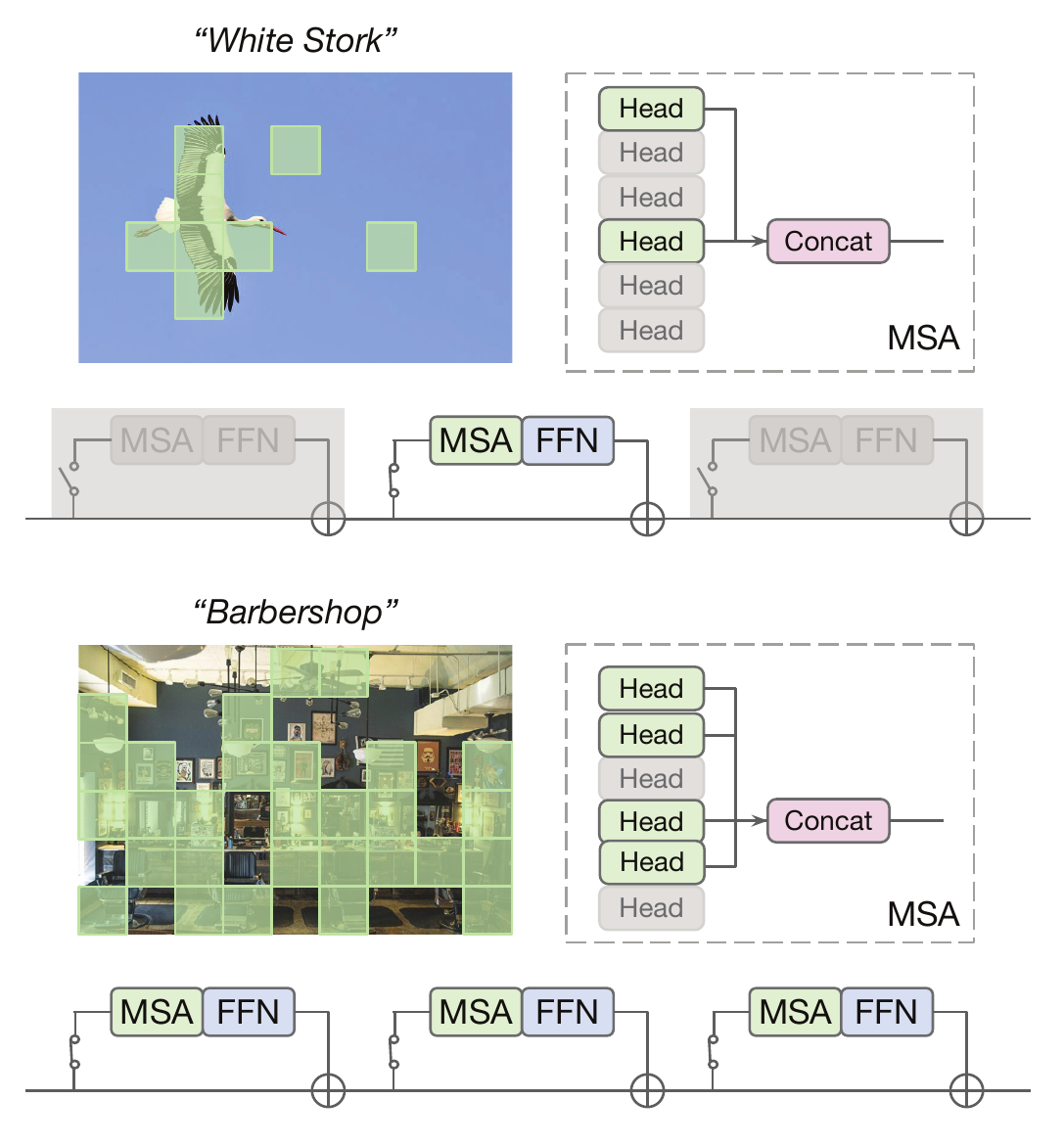}}
    \vspace{-7mm}
    \caption{\textbf{A conceptual overview of our method.} Exploiting the redundancy in vision transformers, AdaViT learns to produce instance-specific usage policies on which patches, self-attention heads and transformer blocks to keep/activate throughout the network for efficient image recognition. Fewer computational resources are allocated for easy samples (top) while more are used for hard samples (bottom), reducing  the overall computational cost with a minimal drop of classification accuracy. Green patches are activated in both figures.}
\vspace{-3mm}
\label{fig:teaser}
\end{figure}

Recent studies on vision transformers~\cite{vit,touvron2021training_deit,t2t,chen2021crossvit} typically adopt the Transformer~\cite{transformer} architecture from NLP with minimal surgery. Taking a sequence of sliced image patches analogous to tokens/words as inputs, the transformer backbone consists of stacked building blocks with two sublayers, \ie a self-attention layer and a feed-forward network. To ensure that the model can attend to information from different representation subspaces jointly, multi-head attention is used in each block instead of a single attention function~\cite{transformer}. While these self-attention-based vision transformers have outperformed CNNs on a multitude of benchmarks like ImageNet~\cite{deng2009imagenet}, the competitive performance does not come for free---the computational cost of the stacked attention blocks with multiple heads is large, which further grows quadratically with the number of patches. 

But are all patches needed to be attended to throughout the network for correctly classifying images? Do we need all the self-attention blocks with multiple heads to \emph{look for} where to attend to and model the underlying dependencies for all different images? After all, large variations exist in images such as object shape, object size, occlusion and background complexity. Intuitively, more patches and self-attention blocks are required for complex images containing cluttered background or occluded objects, which require sufficient contextual information and understanding of the whole image so as to infer their ground-truth classes (\eg the barber shop in Figure~\ref{fig:teaser}), while only a small number of informative patches and attention heads/blocks are enough to classify easy images correctly.

With this in mind, we seek to develop an adaptive computation framework that learns which patches to use and which self-attention heads/blocks to activate on a per-input basis. By doing so, the computational cost of vision transformers can be saved through discarding redundant input patches and backbone network layers for easy samples, and only using full model with all patches for hard and complex samples. This is an orthogonal and complementary direction to recent approaches on efficient vision transformers that focus on designing static network architectures~\cite{liu2021swin,chen2021crossvit,t2t,graham2021levit}.

To this end, we introduce Adaptive Vision Transformer (AdaViT), an end-to-end framework that adaptively determines the usage of patches, heads and layers of vision transformers conditioned on input images for efficient image classification. Our framework learns to derive instance-specific inference strategies on: 1) which patches to keep; 2) which self-attention heads to activate; and 3) which transformer blocks to skip for each image, to improve the inference efficiency with a minimal drop of classification accuracy. In particular, we insert a light-weight multi-head subnetwork (\ie a decision network) to each transformer block of the backbone network, which learns to predict binary decisions on the usage of patch embeddings, self-attention heads and blocks throughout the network. Since binary decisions are non-differentiable, we resort to  Gumbel-Softmax~\cite{maddison2016concrete_gumbel} during training to make the whole framework end-to-end trainable. The decision network is jointly optimized with the transformer backbone with a usage loss that measures the computational cost of the produced usage policies and a normal cross-entropy loss, which incentivizes the network to produce policies that reduce the computational cost while maintaining classification accuracy. The overall \emph{target} computational cost can be controlled by hyperparameter $\gamma \in (0, 1]$ corresponding to the percentage of computational cost of the full model with all patches as input during training, making the framework flexible to suit the need of different computational budgets.

We conduct extensive experiments on ImageNet~\cite{deng2009imagenet} to validate the effectiveness of AdaViT and show that our method is able to improve the inference efficiency of vision transformers by more than $2\times$ with only $0.8\%$ drop of classification accuracy, achieving good trade-offs between efficiency and accuracy when compared with other standard vision transformers and CNNs. In addition, we conduct quantitative and qualitative analyses on the learned usage policies, providing more intuitions and insights on the redundancy in vision transformers. We further show visualizations and demonstrate that AdaViT learns to use more computation for relatively hard samples with complex scenes, and less for easy object-centric samples.

%%%%%%%%% Related Work
\section{Related Work}

\noindent\textbf{Vision Transformers.}
Inspired by its great success in NLP tasks, many recent studies have explored adapting the Transformer~\cite{transformer} architecture to multiple computer vision tasks~\cite{vit,liu2021swin,fan2021multiscale,wang2021end_vistr,ranftl2021vision,xie2021segformer,el2021training,he2021transreid,pan20213d_pointformer,mao2021voxel}. Following ViT~\cite{vit}, a variety of vision transformer variants have been proposed to improve the recognition performance as well as training and inference efficiency. DeiT~\cite{touvron2021training_deit} incorporates distillation strategies to improve training efficiency of vision transformers, outperforming standard CNNs without pretraining on large-scale dataset like JFT~\cite{sun2017revisiting_jft}. Other approaches like T2T-ViT~\cite{t2t}, Swin Transformer~\cite{liu2021swin}, PVT~\cite{wang2021pyramid} and CrossViT~\cite{chen2021crossvit} seek to improve the network architecture of vision transformers. Efforts have also been made to introduce the advantages of 2D CNNs to transformers through using convolutional layers~\cite{li2021localvit,xiao2021early}, hierarchical network structures~\cite{liu2021swin,liu2021videoswin,wang2021pyramid}, multi-scale feature aggregation~\cite{fan2021multiscale,chen2021crossvit} and so on. While obtaining superior performance, the computational cost of vision transformers is still intensive and scales up quickly as the numbers of patches, self-attention heads and transformer blocks increase. 

\noindent\textbf{Efficient Networks.} Extensive studies have been conducted to improve the efficiency of CNNs for vision tasks through designing effective light-weight network architectures like MobileNets~\cite{mobilenets,mobilenetv2,howard2019searching_mobilenetv3}, EfficientNets~\cite{tan2019efficientnet} and ShuffleNets~\cite{zhang2018shufflenet,ma2018shufflenetv2}. To match the inference efficiency of standard CNNs, recent work has also explored developing efficient vision transformer architectures. T2T-ViT~\cite{t2t} proposes to use a deep-narrow structure and a token-to-token module, achieving better accuracy and less computational cost than ViT~\cite{vit}. LeViT~\cite{graham2021levit} and Swin Transformer~\cite{liu2021swin} develop multi-stage network architectures with down-sampling and obtain better inference efficiency. These methods, however, use a fixed network architecture for all input samples regardless of the redundancy in patches and network architecture for easy samples. Our work is orthogonal to this direction and focuses on learning input-specific strategies that adaptively allocate computational resources for saved computation and a minimal drop in accuracy at the same time. 

\noindent\textbf{Adaptive Computation.} Adaptive computation methods exploit the large variations within network inputs as well as the redundancy in network architectures to improve efficiency with instance-specific inference strategies. In particular, existing methods for CNNs have explored altering input resolution~\cite{autofocus,arnet,huanggaoresolution,whenandwhere}, skipping network layers~\cite{andreasadaptive,skipnet,blockdrop,figurnov2017spatially} and channels~\cite{channelgated,runtime}, early exiting with a multi-classifier structure~\cite{icmladaptive,multiscale_densenet,huanggaoimproved}, to name a few. A few attempts have also been made recently to accelerate vision transformers with adaptive inference policies exploiting the redundancy in patches, \ie producing policies on what patch size~\cite{wang2021not} and which patches~\cite{pan2021iared2,rao2021dynamicvit} to use conditioned on input image. In contrast, we exploit the redundancy in the attention mechanism of vision transformer and propose to improve efficiency by adaptively choosing which self-attention heads, transformer blocks and patch embeddings to keep/drop conditioned on the input samples.

\begin{figure*}[!t] \centering
    \resizebox{\linewidth}{!}{\includegraphics[width=\linewidth]{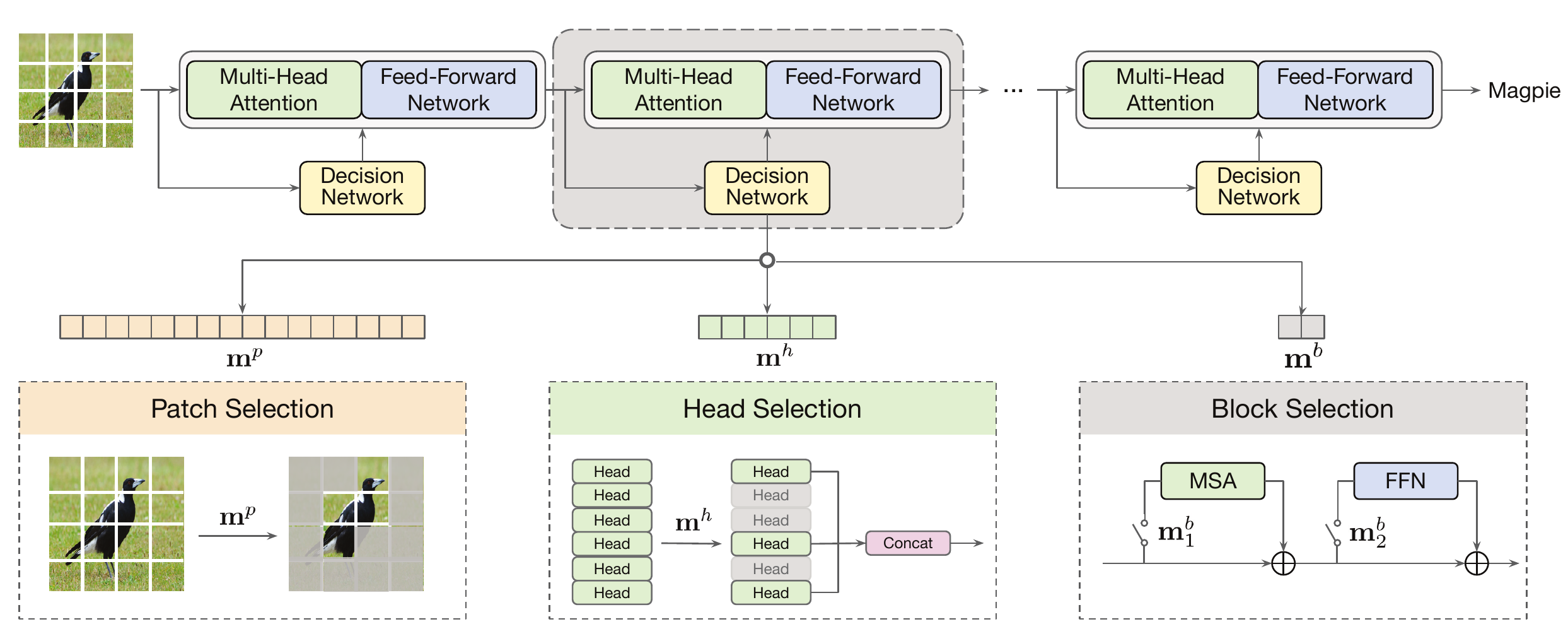}}
    \vspace{-7mm}
    \caption{\textbf{An overview of our approach.} We insert a light-weight decision network before each block of the vision transformer backbone. Given an input image, the decision networks produce usage policies on which \emph{patches}, self-attention \emph{heads} and transformer \emph{blocks} to keep/activate throughout the backbone. These instance-specific usage policies are incentivized to reduce the overall computational cost of vision transformers with minimal drop of accuracy. See texts for more details.}
    \vspace{-4mm}
\label{fig:approach}
\end{figure*}

%%%%%%%%% Approach
\section{Approach}

We propose AdaViT, an end-to-end adaptive computation framework to reduce the computational cost of vision transformers. Given an input image, AdaViT learns to adaptively derive policies on which \emph{patches}, self-attention \emph{heads} and transformer \emph{blocks} to use or activate in the transformer backbone conditioned on the input image, encouraging using less computation while maintaining the classification accuracy. An overview of our method is shown in Figure~\ref{fig:approach}. In this section, we first give a brief introduction of vision transformers in Sec.~\ref{sec:approach_prelim}. We then present our proposed method in Sec.~\ref{sec:approach_adavit} and elaborate the optimization function of the framework in Sec.~\ref{sec:approach_objfunc}. 

\subsection{Preliminaries} \label{sec:approach_prelim}

Vision transformers~\cite{vit,touvron2021training_deit,t2t} for image classification take a sequence of sliced patches from image as input, and model their long-range dependencies with stacked multi-head self-attention layers and feed-forward networks\footnote{In this section we consider the architecture of ViT~\cite{vit}, and extend it to other variants of vision transformers is straightforward.}. Formally, for an input image $\mathcal{I}$, it is first split into a sequence of fixed-size 2D patches $\mathbf{X} = [\mathbf{x}_1, \mathbf{x}_2, ... ,\mathbf{x}_N]$ where $N$ is the number of patches (\eg $N = 14 \times 14$). These raw patches are then mapped into $D$-dimensional patch embeddings $\mathbf{Z} = [\mathbf{z}_1, \mathbf{z}_2, ... ,\mathbf{z}_N]$  with a linear layer. A learnable embedding $\mathbf{z}_{cls}$ termed class token is appended to the sequence of patch embeddings, which serves as the representation of image. Positional embeddings $\mathbf{E}_{pos}$ are also optionally added to patch embeddings to augment them with positional information. To summarize, the input to the first transformer block is:
\begin{align} 
    {\mathbf{Z} = [\mathbf{z}_{cls}; \mathbf{z}_1; \mathbf{z}_2; ... ;\mathbf{z}_N] + \mathbf{E}_{pos}}
\end{align}
where $\mathbf{z} \in \mathbb{R}^{D}$ and $\mathbf{E}_{pos} \in \mathbb{R}^{(N+1) \times D}$ respectively. 

Similar to Transformers~\cite{transformer} in NLP, the backbone network of vision transformers consist of $L$ blocks, each of which consists of a multi-head self-attention layer (MSA) and a feed-forward network (FFN). In particular, a single-head attention is computed as below:
\begin{align}
    \texttt{Attn}(Q, K, V) = \texttt{softmax}(\frac{QK^T}{\sqrt{d_k}})V
\label{eqn:attn}
\end{align}
where $Q, K, V$ are---in a broad sense---query, key and value matrices respectively, and $d_k$ is a scaling factor. For vision transformers, $Q, K, V$ are projected from the same input, \ie patch embeddings. For more effective attention on different representation subspaces, multi-head self-attention concatenates the output from several single-head attentions and projects it with another parameter matrix:
\begin{align}
    \text{head}_{i,l} &= \texttt{Attn}(\mathbf{Z}_{l}\mathbf{W}_{i,l}^Q, \ \mathbf{Z}_{l}\mathbf{W}_{i,l}^K, \ \mathbf{Z}_{l}\mathbf{W}_{i,l}^V) \\
    \texttt{MSA}(\mathbf{Z}_{l}) &= \texttt{Concat}(\text{head}_{1,l}, ..., \text{head}_{H,l})\mathbf{W}_l^O,
\end{align}
where $\mathbf{W}_{i,l}^Q, \mathbf{W}_{i,l}^K, \mathbf{W}_{i,l}^V, \mathbf{W}_l^O$ are the parameter matrices in the $i$-th attention head of the $l$-th transformer block, and $\mathbf{Z}_{l}$ denotes the input at the $l$-th block. The output from MSA is then fed into FFN, a two-layer MLP, and produce the output of the transformer block $\mathbf{Z}_{l+1}$. Residual connections are also applied on both MSA and FFN as follows:
\begin{align}
    \mathbf{Z}'_{l} = \texttt{MSA}(\mathbf{Z}_{l}) + \mathbf{Z}_{l}, \quad \mathbf{Z}_{l+1} = \texttt{FFN}(\mathbf{Z}'_{l}) + \mathbf{Z}'_{l}
\label{eqn:block}
\end{align}
The final prediction is produced by a linear layer taking the class token from last transformer block ($\mathbf{Z}^0_{L}$) as inputs.

\subsection{Adaptive Vision Transformer} \label{sec:approach_adavit}

While large vision transformer models have achieved superior image classification performance, the computational cost grows quickly as we increase the numbers of patches, attention heads and transformer blocks to obtain higher accuracies. In addition, a computationally expensive one-size-fit-all network is often an overkill for many easy samples.
To remedy this, AdaViT learns to adaptively choose 1) which patch embeddings to use; 2) which self-attention heads in MSA to activate; and 3) which transformer block to skip---on a per-input basis---to improve the inference efficiency of vision transformers. We achieve this by inserting a light-weight decision network before each of the transformer blocks, and it is trained to produce the three sets of usage policies for this block.

\vspace{1mm}
\noindent\textbf{Decision Network.} The decision network at $l$-th block consists of three linear layers with parameters $\mathbf{W}_l = \{ \mathbf{W}^p_l, \mathbf{W}^h_l, \mathbf{W}^b_l \}$ to produce computation usage policies for \emph{patch selection}, attention \emph{head selection} and transformer \emph{block selection} respectively. Formally, given the input to $l$-th block $\mathbf{Z}_{l}$, the usage policy matrices for this block is computed as follows:
\begin{align}
    (\mathbf{m}^p_l, \mathbf{m}^h_l, \mathbf{m}^b_l) = & (\mathbf{W}^p_l, \mathbf{W}^h_l, \mathbf{W}^b_l) \mathbf{Z}_{l}  \nonumber \\
    \text{\emph{s.t.}} & \ \  \mathbf{m}^p_l \in \mathbb{R}^{N}, \ \mathbf{m}^h_l \in \mathbb{R}^{H}, \ \mathbf{m}^b_l \in \mathbb{R} 
\label{eqn:policy_prob}
\end{align}
where $N$ and $H$ denote the numbers of patches and self-attention heads in a transformer block, and $l \in [1, L]$. Each entry of $\mathbf{m}^p_l$, $\mathbf{m}^h_l$ and $\mathbf{m}^b_l$ is further passed to a \texttt{sigmoid} function, indicating the probability of keeping the corresponding patch, attention head and transformer block respectively. The $l$-th decision network shares the output from previous $l-1$ transformer blocks, making the framework more efficient than using a standalone decision network. 

As the decisions are binary, the action of keeping / discarding can be selected by simply applying a threshold on the entries during inference. However, deriving the optimal thresholds for different samples is challenging. To this end, we define random variables $\mathbf{M}^p_l$, $\mathbf{M}^h_l$, $\mathbf{M}^b_l$ to make decisions by sampling from $\mathbf{m}^p_l$, $\mathbf{m}^h_l$ and $\mathbf{m}^b_l$. For example, the $j$-th patch embedding in $l$-th block is kept when $\mathbf{M}^p_{l,j}=1$, and dropped when $\mathbf{M}^p_{l,j}=0$. We relax the sampling process with Gumbel-Softmax trick~\cite{maddison2016concrete_gumbel} to make it differentiable during training, which will be further elaborated in Sec.~\ref{sec:approach_objfunc}.

\vspace{1mm}
\noindent\textbf{Patch Selection.} For the input to each transformer block, we aim at keeping only the most informative patch embeddings and discard the rest to speedup inference. More formally, for $l$-th block, the patches are removed from the input to this block if the corresponding entries in $\mathbf{M}^p_{i}$ equal to $0$:
\begin{align}
    \mathbf{Z}_{l} = [\mathbf{z}_{l, cls} ; \mathbf{M}^p_{l,1} \mathbf{z}_1 ; ... ; \mathbf{M}^p_{l,N} \mathbf{z}_N]
\end{align}
The class token $\mathbf{z}_{l, cls}$ is always kept since it is used as representation of the whole image.

\vspace{1mm}
\noindent\textbf{Head Selection.} Multi-head self attention enables the model to attend to different subspaces of the representation jointly~\cite{transformer} and is adopted in most, if not all, vision transformer variants~\cite{vit,touvron2021training_deit,t2t,chen2021crossvit,liu2021swin}. Such a multi-head design is crucial to model the underlying long-range dependencies in images especially those with complex scenes and cluttered background, but fewer attention heads could arguably suffice to look for where to attend to in easy images. With this in mind, we explore dropping attention heads adaptively conditioned on input image for faster inference. Similar to patch selection, the decision of activating or deactivating certain attention head is determined by the corresponding entry in $\mathbf{M}^h_l$. The ``deactivation'' of an attention head can be instantiated in different ways. In our framework, we explore two methods for head selection, namely partial deactivation and full deactivation. 

For \emph{partial deactivation}, the softmax output in attention as in Eqn.~\ref{eqn:attn} is replaced with predefined ones like an $(N+1, N+1)$ identity matrix $\mathbb{1}$, such that the cost of computing attention map is saved. The attention in $i$-th head of $l$-th block is then computed as:
\begin{align}
\texttt{Attn}(Q, K, V)_{l, i} = 
\begin{cases}
\texttt{softmax}(\frac{QK^T}{\sqrt{d_k}}) \cdot V & \text{if} \ \mathbf{M}^h_{l, i} = 1 \\
\mathbb{1} \cdot V & \text{if} \ \mathbf{M}^h_{l, i} = 0 \\
\end{cases}
\end{align}

For \emph{full deactivation}, the entire head is removed from the multi-head self attention layer, and the embedding size of the output from MSA is reduced correspondingly:
\begin{align}
\texttt{MSA}(\mathbf{Z}_{l})_{l, i} &= \texttt{Concat}([\text{head}_{l, i:1 \veryshortarrow H} \ \ \text{if} \ \ \mathbf{M}^h_{l, i}=1])\mathbf{W}_l^{O'} 
\end{align}
In practice, full deactivation saves more computation compared with partial deactivation when same percentage of heads are deactivated , yet is likely to incur more classification errors as the embedding size is manipulated on-the-fly.  

\vspace{1mm}
\noindent\textbf{Block Selection.} In addition to patch selection and head selection, a transformer block can also be favourably skipped entirely when it is redundant, by virtue of the residual connections throughout the network. To increase the flexibility of layer skipping, we increase the dimension of block usage policy matrix $\mathbf{m}^b_l$ from $1$ to $2$, enabling the two sublayers (MSA and FFN) in each transformer block to be controlled individually. Eqn.~\ref{eqn:block} then becomes:
\begin{align}
    \mathbf{Z}'_{l} = \mathbf{M}^b_{l, 0} \cdot \texttt{MSA}(\mathbf{Z}_{l}) + \mathbf{Z}_{l} \nonumber \\
    \mathbf{Z}_{l+1} = \mathbf{M}^b_{l, 1} \cdot \texttt{FFN}(\mathbf{Z}'_{l}) + \mathbf{Z}'_{l}
\end{align}

In summary, given the input of each transformer block, the decision network produces the usage policies for this block, and then the input is forwarded through the block with the decisions applied. Finally, the classification prediction from the last layer and the decisions for all blocks $\mathbf{M} = \{\mathbf{M}^p_l$, $\mathbf{M}^h_l$, $\mathbf{M}^b_l, \text{for} \ l:1 \veryshortarrow L \}$ are obtained. 

\begin{table*}[!t]  \centering
    \resizebox{0.85\linewidth}{!}{\begin{tabular}{l*{8}c}
    \toprule
    Method && Top-1 Acc (\%) & FLOPs (G) & Image Size && \#\,Patch & \#\,Head & \#\,Block \\
    \cmidrule{1-1} \cmidrule{3-5} \cmidrule{7-9} 
    ResNet-50*~\cite{resnet,t2t} && 79.1 & 4.1 & 224$\times$224 && - & - & - \\
    ResNet-101*~\cite{resnet,t2t} && 79.9 & 7.9 & 224$\times$224 && - & - & - \\
    \cmidrule{1-1} \cmidrule{3-5} \cmidrule{7-9} 
    ViT-S/16~\cite{vit} && 78.1 & 10.1 & 224$\times$224 && 196 & 12 & 8 \\
    DeiT-S~\cite{touvron2021training_deit} && 79.9 & 4.6 & 224$\times$224 && 196 & 6 & 12 \\
    PVT-Small~\cite{wang2021pyramid} && 79.8 & 3.8 & 224$\times$224 && - & - & 15 \\
    Swin-T~\cite{liu2021swin} && 81.3 & 4.5 & 224$\times$224 && - & - & 12 \\
    T2T-ViT-19~\cite{t2t} && 81.9 & 8.5 & 224$\times$224 && 196 & 7 & 19 \\
    CrossViT-15~\cite{chen2021crossvit} && 81.5 & 5.8 & 224$\times$224 && 196 &  6 & 15 \\
    LocalViT-S~\cite{li2021localvit} && 80.8 & 4.6 & 224$\times$224 && 196 & 6 & 12 \\
    \cmidrule{1-1} \cmidrule{3-5} \cmidrule{7-9} 
    Baseline \emph{Upperbound} && 81.9 & 8.5 & 224$\times$224 && 196 & 7 & 19 \\
    Baseline \emph{Random} && 33.0 & 4.0 & 224$\times$224 && $\sim$ 118 & $\sim$ 5.6 & $\sim$ 16.2 \\
    Baseline \emph{Random+} && 71.5  & 3.9 & 224$\times$224 &&  $\sim$ 121 & $\sim$ 5.6 & $\sim$ 16.2 \\
    \textbf{AdaViT (Ours)} && 81.1 & 3.9 & 224$\times$224 && $\sim$ 95 & $\sim$ 4.5 & $\sim$ 15.5 \\
    \bottomrule
    \end{tabular}}
\vspace{-1mm}
\caption{\textbf{Main Results.} We compare AdaViT with various standard CNNs and vision transformers, as well as baselines including \emph{Upperbound}, \emph{Random} and \emph{Random+}. * denotes training ResNets with our recipe following~\cite{t2t}.}
\vspace{-5mm}
\label{table:main}
\end{table*}

\subsection{Objective Function} \label{sec:approach_objfunc}

Since our goal is to reduce the overall computational cost of vision transformers with a minimal drop in accuracy, the objective function of AdaViT is designed to incentivize correct classification and less computation at the same time. In particular, a usage loss and a cross-entropy loss are used to jointly optimize the framework. Given an input image $I$ with a label $\mathbf{y}$, the final prediction is produced by the transformer $\mathbf{F}$ with parameters $\bm \theta$, and the cross-entropy loss is computed as follows:
\begin{align}
    L_{ce} = -\mathbf{y} \texttt{log}(\mathbf{F}(I; \bm{\theta}))
\end{align}

While the binary decisions on whether to keep/discard a patch/head/block can be readily obtained through applying a threshold during inference, determining the optimal thresholds is challenging. In addition, such an operation is not differentiable during training and thus makes the optimization of decision network challenging. A common solution is to resort to reinforcement learning and optimize the network with policy gradient methods~\cite{policygradient}, yet it can be slow to converge due to the large variance that scales with the dimension of discrete variables~\cite{policygradient,maddison2016concrete_gumbel}. To this end, we use the Gumbel-Softmax trick~\cite{maddison2016concrete_gumbel} to relax the sampling process and make it differentiable. Formally, the decision at $i$-th entry of $\mathbf{m}$ is derived in the following way:
\begin{align}
    \mathbf{M}_{i,k} = \frac{\texttt{exp}(\texttt{log} (\mathbf{m}_{i, k} + G_{i, k}) / \tau)}{\sum^K_{j=1} \texttt{exp}(\texttt{log} (\mathbf{m}_{i, j} + G_{i, j}) / \tau)} \nonumber \\
    \text{for} \ k = 1, 2, ..., K
\end{align}
where $K$ is the total number of categories ($K=2$ for binary decision in our case), and $G_i = -\texttt{log} (-\texttt{log} (U_i))$ is the Gumbel distribution in which $U_i$ is sampled from $\texttt{Uniform}(0, 1)$, an i.i.d uniform distribution. Temperature $\tau$ is used to control the smoothness of $\mathbf{M}_i$. 

To encourage reducing the overall computational cost, we devise the usage loss as follows:
\begin{align}
    L_{usage} &= (\frac{1}{D_p} \sum_{d=1}^{D_p} \mathbf{M}^p_d - \gamma_p)^2 + (\frac{1}{D_h} \sum_{d=1}^{D_h} \mathbf{M}^h_d - \gamma_h)^2 \nonumber \\ 
    &+ (\frac{1}{D_b} \sum_{d=1}^{D_b} \mathbf{M}^b_d - \gamma_b)^2 \nonumber \\
    \text{where} \ & D_p = L \times N, \ D_h = L \times H, \ D_b = L \times 2
\end{align}
Here $D_p, D_h, D_b$ denote the sizes of flattened probability vectors from the decision network for patch/head/block selection, \ie the total numbers of patches, heads and blocks of the entire transformer respectively. The hyperparameters $\gamma_p, \gamma_h, \gamma_b \in (0, 1]$ indicate target computation budgets in terms of the percentage of patches/heads/blocks to keep. 
\begin{align}
    \min\limits_{\bm{\theta}, \mathbf{W}} \ L = L_{ce} + L_{usage}
\label{eqn:loss_final}
\end{align}
Finally, the two loss functions are combined and minimized in an end-to-end manner as in Eqn.~\ref{eqn:loss_final}.

%%%%%%%%% Experiments
\section{Experiment}

\subsection{Experimental Setup}

\noindent\textbf{Dataset and evaluation metrics.} We conduct experiments on ImageNet~\cite{deng2009imagenet} with $\sim$1.2M images for training and 50K images for validation, and report the Top-1 classification accuracy. To evaluate model efficiency, we report the number of giga floating-point operations (GFLOPs) per image. 

\noindent\textbf{Implementation details.} We use T2T-ViT~\cite{t2t} as the transformer backbone due to its superior performance on ImageNet with a moderate computational cost. The backbone consists of $L=19$ blocks and $H=7$ heads in each MSA layer, and the number of tokens $N=196$. The decision network is attached to each transformer block starting from $2$-nd block. For head selection, we use the \emph{full deactivation} method if not mentioned otherwise. We initialize the transformer backbone of AdaViT with the pretrained weights released in the official implementation of~\cite{t2t}. We will release the code. 

We use 8 GPUs with a batch size 512 for training. The model is trained with a learning rate $0.0005$, a weight decay $0.065$ and a cosine learning rate schedule for $150$ epochs following~\cite{t2t}. AdamW~\cite{loshchilov2017decoupled_adamw} is used as the optimizer. For all the experiments, we set the input size to $224 \times 224$. Temperature $\tau$ in Gumbel-Softmax is set to $5.0$. The choices of $\gamma_p, \gamma_h, \gamma_b$ vary flexibly for different desired trade-offs between classification accuracy and computational cost. 

\subsection{Main Results}

We first evaluate the overall performance of AdaViT in terms of classification accuracy and efficiency, and report the results in Table~\ref{table:main}. Besides standard CNN and transformer architectures such as ResNets~\cite{resnet}, ViT~\cite{vit}, DeiT~\cite{touvron2021training_deit}, T2T-ViT~\cite{t2t} and so on, we also compare our method with the following baseline methods:

\begin{itemize} 
    \vspace{-2mm}
    \item \emph{Upperbound}: The original pretrained vision transformer model, with all patch embeddings kept as input and all self-attention heads and transformer blocks activated. This serves as an ``upperbound'' of our method regarding classification accuracy. 
    \vspace{-2mm}
    \item \emph{Random}: Given the usage policies produced by AdaVit, we generate random policies on patch selection, head selection and block selection that use similar computational cost and apply them to the pretrained models to validate the effectiveness of learned policies. 
    \vspace{-2mm}
    \item \emph{Random+}: The pretrained models are further finetuned with the random policies applied, in order to adapt to the varied input distribution and network architecture incurred by the random policies. 
    \vspace{-2mm}
\end{itemize}

\begin{figure}[!b] \centering
\vspace{-3mm}
    \resizebox{0.95\linewidth}{!}{\includegraphics[width=\linewidth]{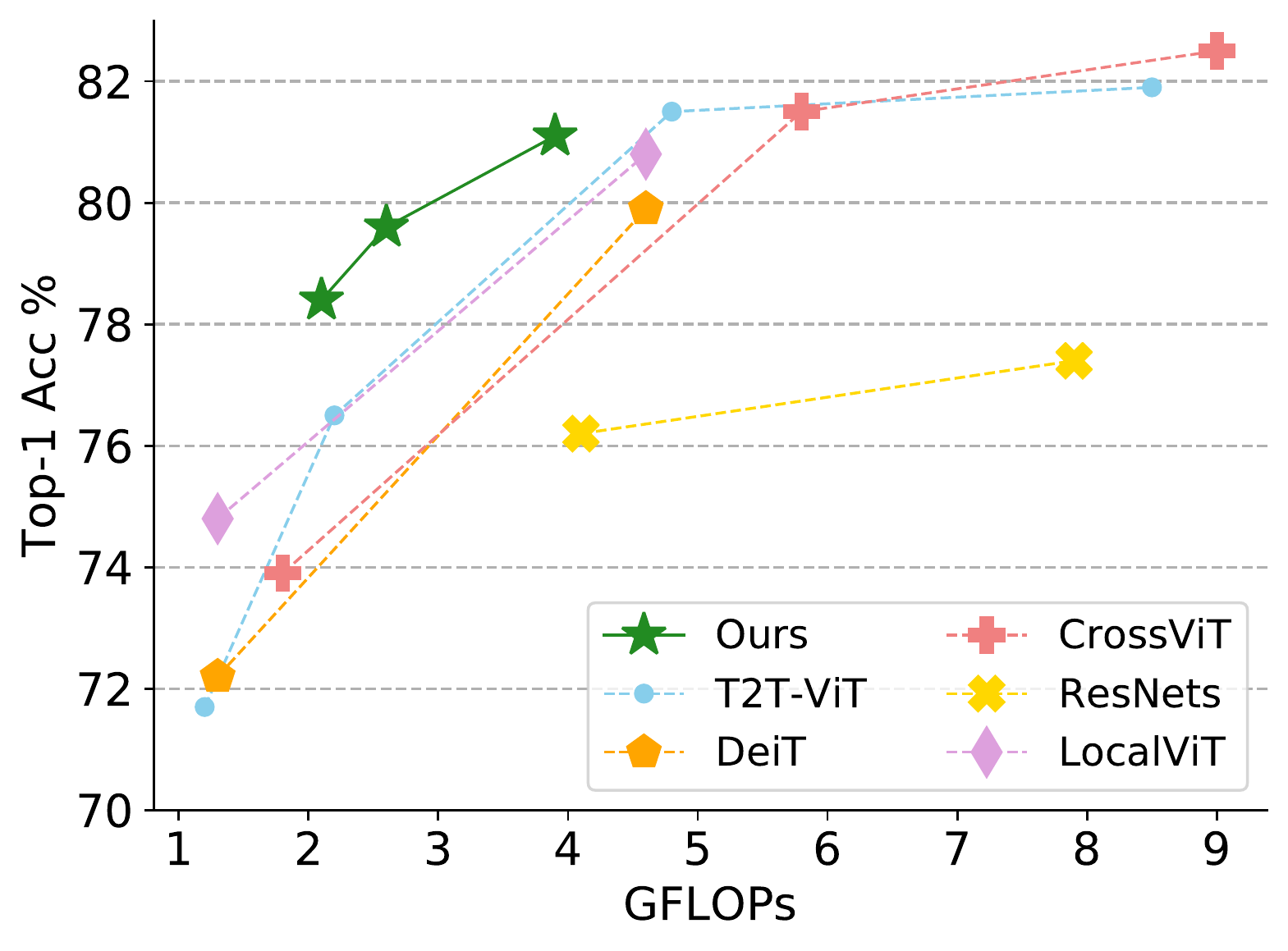}}
    \vspace{-3mm}
    \caption{\textbf{Tradeoff between efficiency and accuracy.} AdaViT obtains good efficiency/accuracy tradeoffs compared with other static vision transformers.}
\label{fig:main_tradeoff}
\end{figure}

\begin{figure*}[!t] \centering
    \resizebox{\linewidth}{!}{\includegraphics[width=\linewidth]{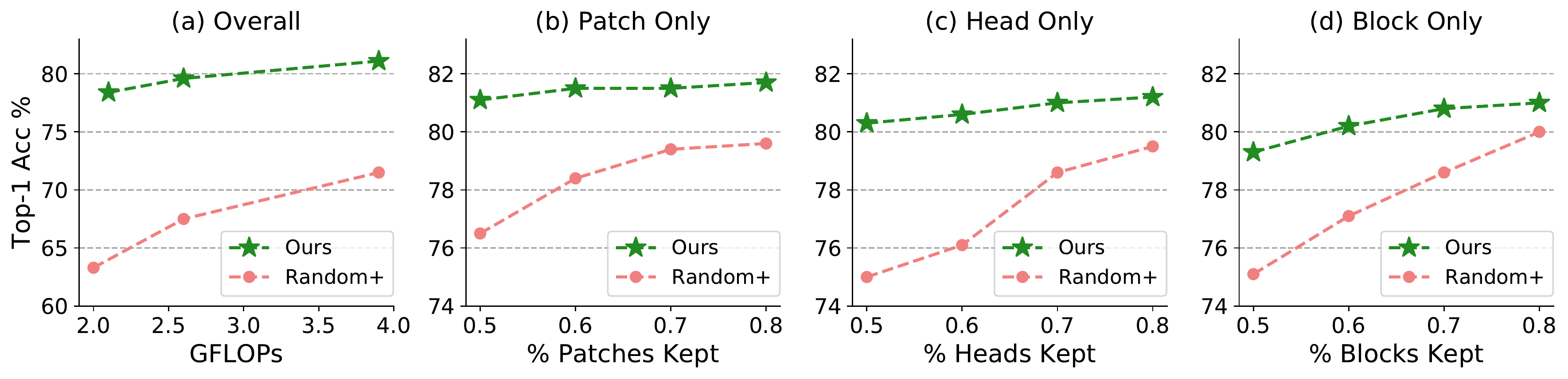}}
    \vspace{-6mm}
    \caption{\textbf{Effectiveness of each component.} Efficiency/Accuracy tradeoffs of AdaViT with (a) all three selection methods; (b) patch selection; (c) head selection; (d) block selection and their \emph{Random+} counterparts.}
\vspace{-3mm}
\label{fig:ab_components}
\end{figure*}

As shown in Table~\ref{table:main}, AdaViT is able to obtain good efficiency improvement with only a small drop on classification accuracy. Specifically, AdaViT obtains $81.1\%$ Top-1 accuracy requiring $3.9$ GFLOPs per image during inference, achieving more than $2\times$ efficiency than the original T2T-ViT model with only $\sim 0.8\%$ drop of accuracy. Compared with standard ResNets~\cite{resnet} and vision transformers that use a similar backbone architecture of ours~\cite{vit,touvron2021training_deit,t2t,chen2021crossvit}, AdaViT obtains better classification performance with less computational cost, achieving a good efficiency/accuracy trade-off as further shown in Figure~\ref{fig:main_tradeoff}. It is also worth pointing out that compared with vision transformer variants~\cite{liu2021swin,wang2021pyramid} which resort to advanced design choices like multi-scale feature pyramid and hierarchical downsampling, our method still obtains comparable or better accuracy under similar computational cost.  

When using a similar computation budget, AdaViT outperforms \emph{random} and \emph{random+} baselines by clear margins. Specifically, Ada-ViT with T2T-ViT as the backbone network obtains $48.1\%$ and $9.6\%$ higher accuracy than \emph{random} and \emph{random+} respectively at a similar cost of $3.9$ GFLOPs per image, demonstrating that the usage policies learned by AdaViT can effectively maintain classification accuracy and reduce computational cost at the same time.  

\noindent\textbf{AdaViT with different computational budgets.} AdaViT is designed to accommodate the need of different computational budgets flexibly by varying the hyperparameters $\gamma_p, \gamma_h$ and $\gamma_b$ as discussed in Section~\ref{sec:approach_adavit}. As demonstrated in Figure~\ref{fig:ab_components}(a), AdaViT is able to cover a wide range of tradeoffs between efficiency and accuracy, and outperforms \emph{Random+} baselines by a large margin. 

\begin{table}[!b]  \centering
    \resizebox{0.8\linewidth}{!}{\begin{tabular}{*{5}c}
    \toprule
    \makecell{Random \\ Patch} & \makecell{Random \\ Head} & \makecell{Random \\ Block} && \makecell{Top-1 \\ Accuracy} \\
    \cmidrule{1-3} \cmidrule{5-5}
    \checkmark & & && 49.2 \\ 
    & \checkmark & && 57.4 \\ 
    & & \checkmark && 64.7 \\ 
    \cmidrule{1-3} \cmidrule{5-5}
    \multicolumn{3}{c}{Full AdaViT} && 81.1 \\
    \bottomrule
    \end{tabular}}
    \vspace{-1mm}
    \caption{\textbf{Effectiveness of learned usage policies.} We replace each set of policies with randomly generated policies and compare with our method in its entirety.}
\label{table:ab_component}
\end{table}

\begin{table}[!b]  \centering
    \resizebox{0.85\linewidth}{!}{\begin{tabular}{*{5}c}
    \toprule
    Method && Top-1 Acc & \% Head & GFLOPs  \\
    \cmidrule{1-1} \cmidrule{3-5}
    Upperbound && 81.9 & 100\% & 8.5 \\
    \cmidrule{1-1} \cmidrule{3-5}
    Partial && 81.7 & 50\% & 6.9 \\
    Full && 80.3 & 50\% & 5.1 \\
    Full && 80.8 & 60\% & 5.8 \\
    Full && 81.1 & 70\% & 6.6 \\
    \bottomrule
    \end{tabular}}
    \vspace{-1mm}
    \caption{\textbf{Partial \vs Full deactivation for head selection.}}
\label{table:ab_head}
\end{table}

\subsection{Ablation Study}

\noindent\textbf{Effectiveness of learned usage policies.} 
Here we validate that each of the three sets of learned usage policies is able to effectively maintain the classification accuracy while reducing the computational cost of vision transformers. For this purpose, we replace the learned usage policies with randomly generated policies that cost similar computational resources and report the results in Table~\ref{table:ab_component}. As shown in Table~\ref{table:ab_component}, changing any set of learned policies to a random one results in a drop of accuracy by a clear margin. Compared with random patch/head/block selection, AdaViT obtains $31.9\%/23.7\%/16.4\%$ higher accuracy under similar computational budget. This confirms the effectiveness of each learned usage policy. 

\noindent\textbf{Ablation of individual components.} Having demonstrated the effectiveness of the jointly learned usage policies for patch, head and block selection, we now evaluate the performance when only one of the three selection methods is used. It is arguable that part of the performance gap in Table~\ref{table:ab_component} results from the change of input/feature distribution when random policies are applied, and thus we compare each component with its further finetuned \emph{Random+} counterparts. For faster training and evaluation, we train these models for $100$ epochs. As shown in Figure~\ref{fig:ab_components}(b-d), our method with only patch/head/block selection is also able to cover a wide range of accuracy/efficiency tradeoffs and outperforms \emph{Random+} baselines by a clear margin, confirming the effectiveness of each component. 

\noindent\textbf{Partial \vs Full deactivation for head selection.} As discussed in Sec.~\ref{sec:approach_adavit}, we propose two methods to deactivate a head in the multi-head self-attention layer, namely partial deactivation and full deactivation. We now analyze their effectiveness on improving the efficiency of vision transformers. As demonstrated in Table~\ref{table:ab_head}, when deactivating the same percentage (\ie $50\%$) of self-attention heads within the backbone, partial deactivation is able to obtain much higher accuracy than full deactivation ($81.7\%$ \vs $80.3\%$), but also incurs higher computational cost ($6.9$ \vs $5.1$ GFLOPs). This is intuitive since partial deactivation only skips the computation of attention maps before \texttt{Softmax}, while full deactivation removes the entire head and its output to the FFN. As the number of heads increases, full deactivation obtains better accuracy gradually. In practice these different head selection methods provide more flexible options to suit different computational budgets. 

\begin{figure}[!t] \centering
    \resizebox{\linewidth}{!}{\includegraphics[width=\linewidth]{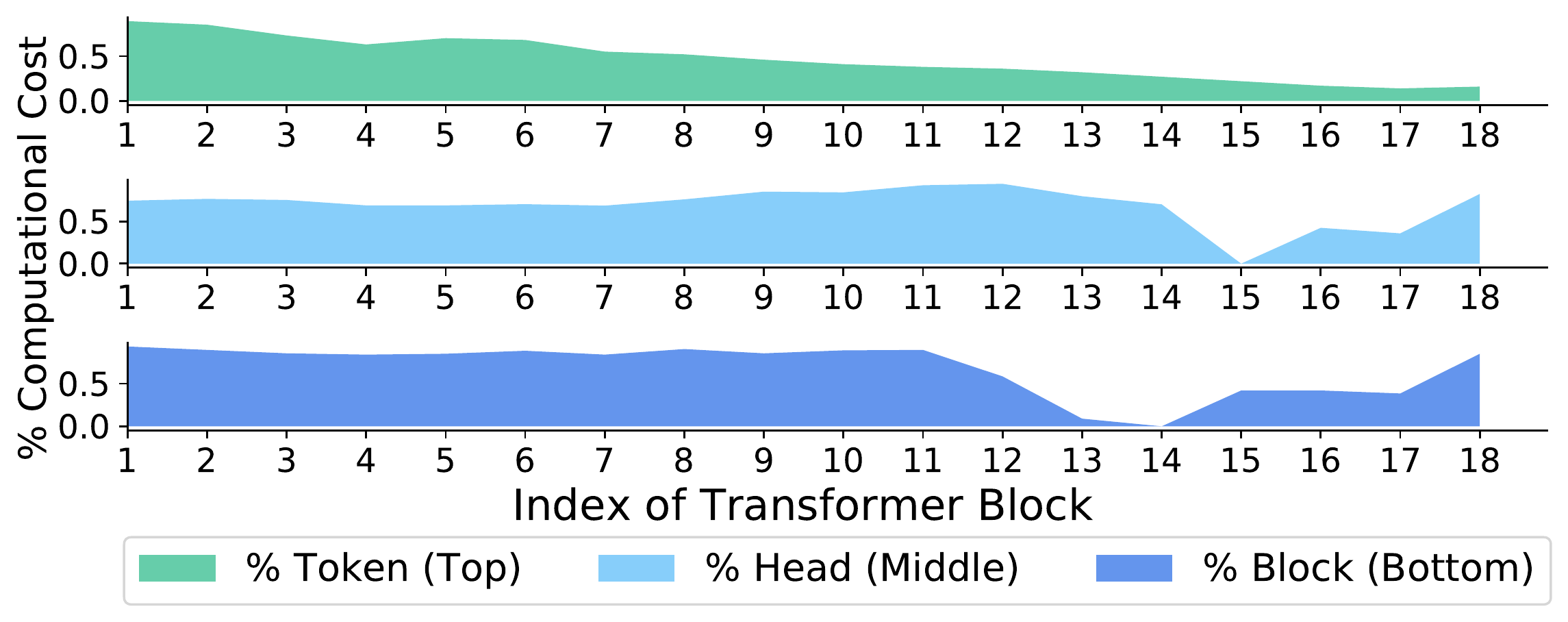}}
    \vspace{-6mm}
    \caption{\textbf{Computational cost throughout the network.} The percentages of kept/activated patches (\textbf{top}), heads (\textbf{middle}) and blocks (\textbf{bottom}) throughout the backbone are reported.}
\label{fig:usage_thru_network}
\end{figure}

\begin{figure}[!t] \centering
    \resizebox{\linewidth}{!}{\includegraphics[width=\linewidth]{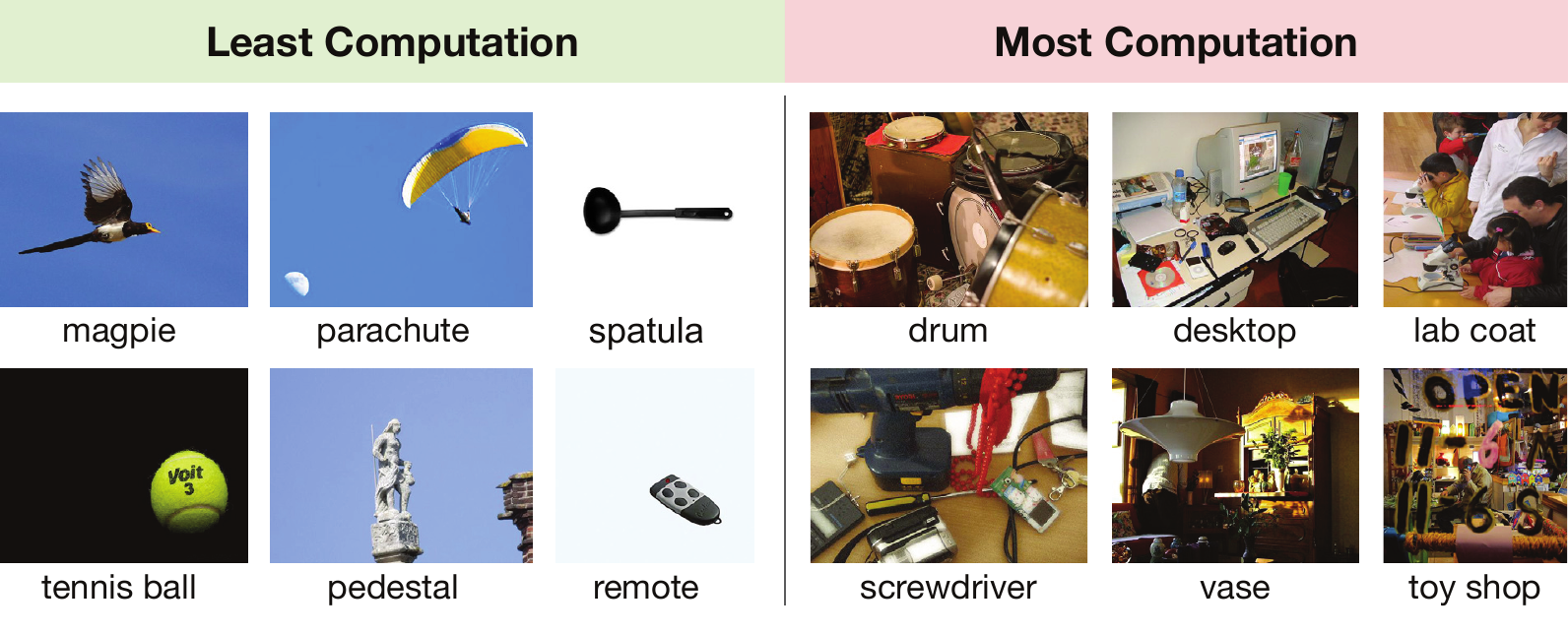}}
    \vspace{-7mm}
    \caption{\textbf{Qualitative results.} Images allocated with the least and the most computational resources by AdaViT are shown.}
\label{fig:qual}
\end{figure}

\subsection{Analysis}

\begin{figure}[!t] \centering
    \resizebox{\linewidth}{!}{\includegraphics[width=\linewidth]{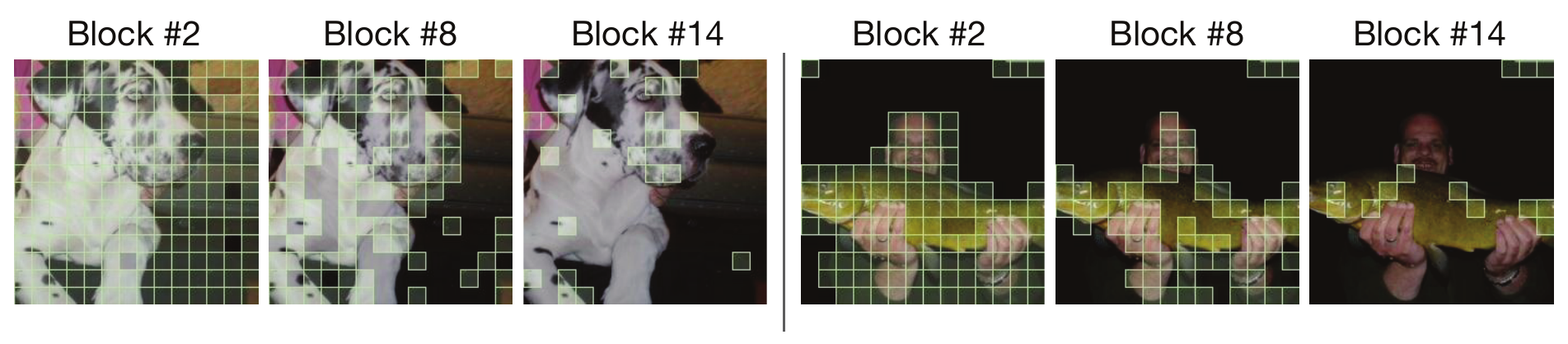}}
    \vspace{-7mm}
    \caption{\textbf{Selected patches at different blocks.} Green color denotes that the patches are kept.}
\label{fig:patch_visualize}
\end{figure}

\noindent\textbf{Computational saving throughout the network.} AdaViT exploits the redundancy of computation to improve the efficiency of vision transformers. To better understand such redundancy, we collect the usage policies on patch/head/block selection predicted by our method on the validation set and show the distribution of computational cost (\ie percentage of patches/heads/blocks kept) throughout the backbone network. As shown in Figure~\ref{fig:usage_thru_network}, AdaViT tends to allocate more computation in earlier stages of the network. In particular, for patch selection, the average number of kept patches in each transformer block gradually decrease until the final output layer. This is intuitive since the patches keep aggregating information from all other patches in the stacked self-attention layers, and a few informative patches near the output layer would suffice to represent the whole input image for correct classification. As visualized in Figure~\ref{fig:patch_visualize}, the number of selected patches gradually decreases with a focus on the discriminative part of the images.

For head selection and block selection, the patterns are a bit different from token selection, where relatively more computation is kept in the last few blocks. We hypothesize that the last few layers in the backbone are more responsible for the final prediction and thus are kept more often.

\begin{figure}[!t] \centering
    \resizebox{\linewidth}{!}{\includegraphics[width=\linewidth]{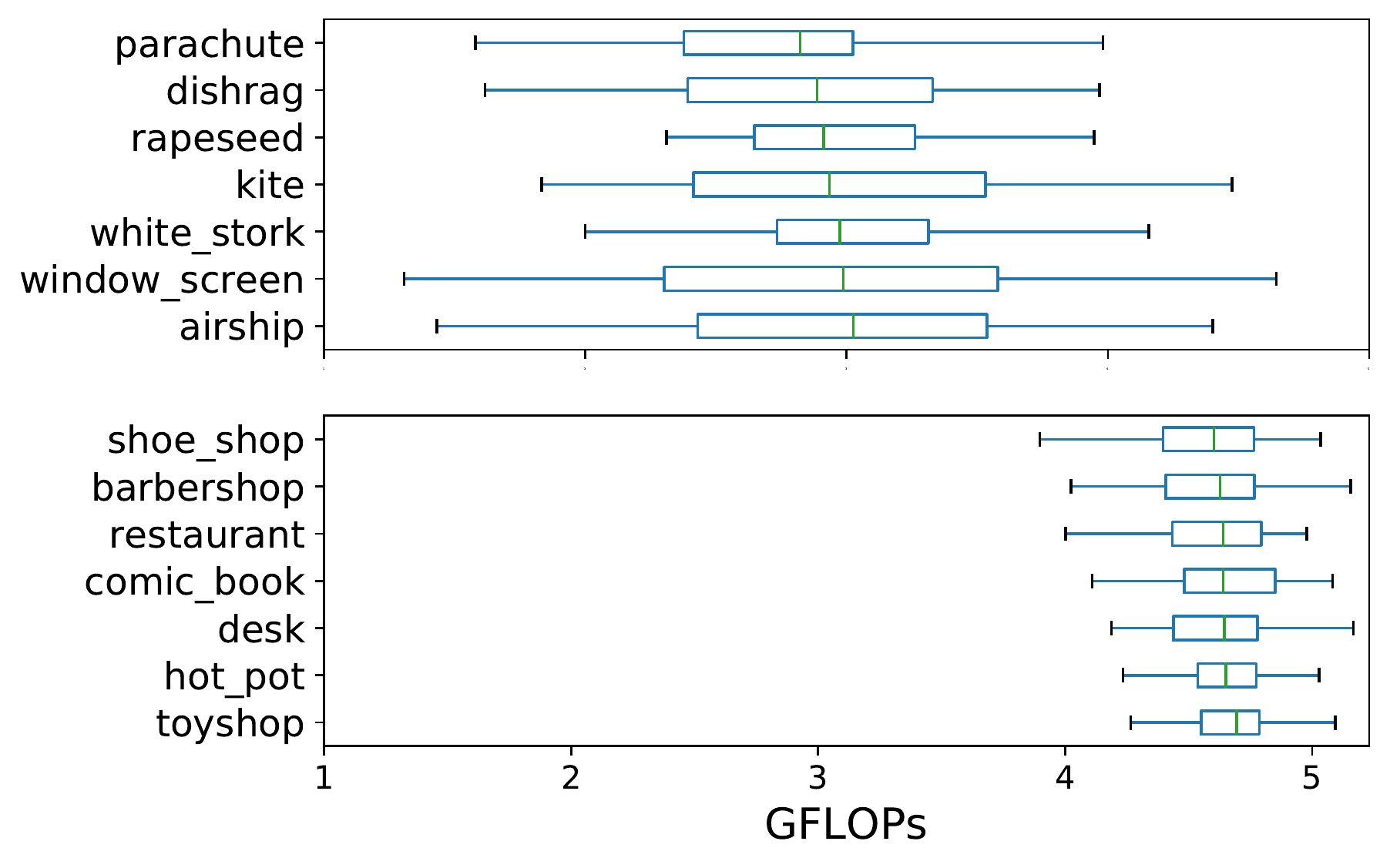}}
    \vspace{-5mm}
    \caption{\textbf{Distribution of allocated computational resources} for classes using the least (\textbf{top}) and the most (\textbf{bottom}) computation.}
\label{fig:ana_boxplot}
\end{figure}

\noindent\textbf{Learned usage policies for different classes.} We further analyze the distribution of learned usage policies for different classes. In Figure~\ref{fig:ana_boxplot}, we show the box plot of several classes that are allocated the most/least computational resources. As can be seen, our method learns to allocate more computation for difficult classes with complex scenes such as ``shoe shop'', ``barber shop'', ``toyshop'' but uses less computation for relatively easy and object-centric classes like ``parachute'' and ``kite''. 

\noindent\textbf{Qualitative Results.} Images allocated with the least and the most computation by our method are shown in Figure~\ref{fig:qual}. It can be seen that object-centric images with simple background (like the parachute and the tennis ball) tend to use less computation, while hard samples with clutter background (\eg the drum and the toy shop) are allocated more. 

\noindent{\textbf{Limitation}. One potential limitation is that there is still a small drop of accuracy when comparing our method with the
\emph{Upperbound} baseline, which we believe would be further addressed in future work.}

\begin{figure*}[!t] \centering
    \resizebox{\linewidth}{!}{\includegraphics[width=1.0\linewidth]{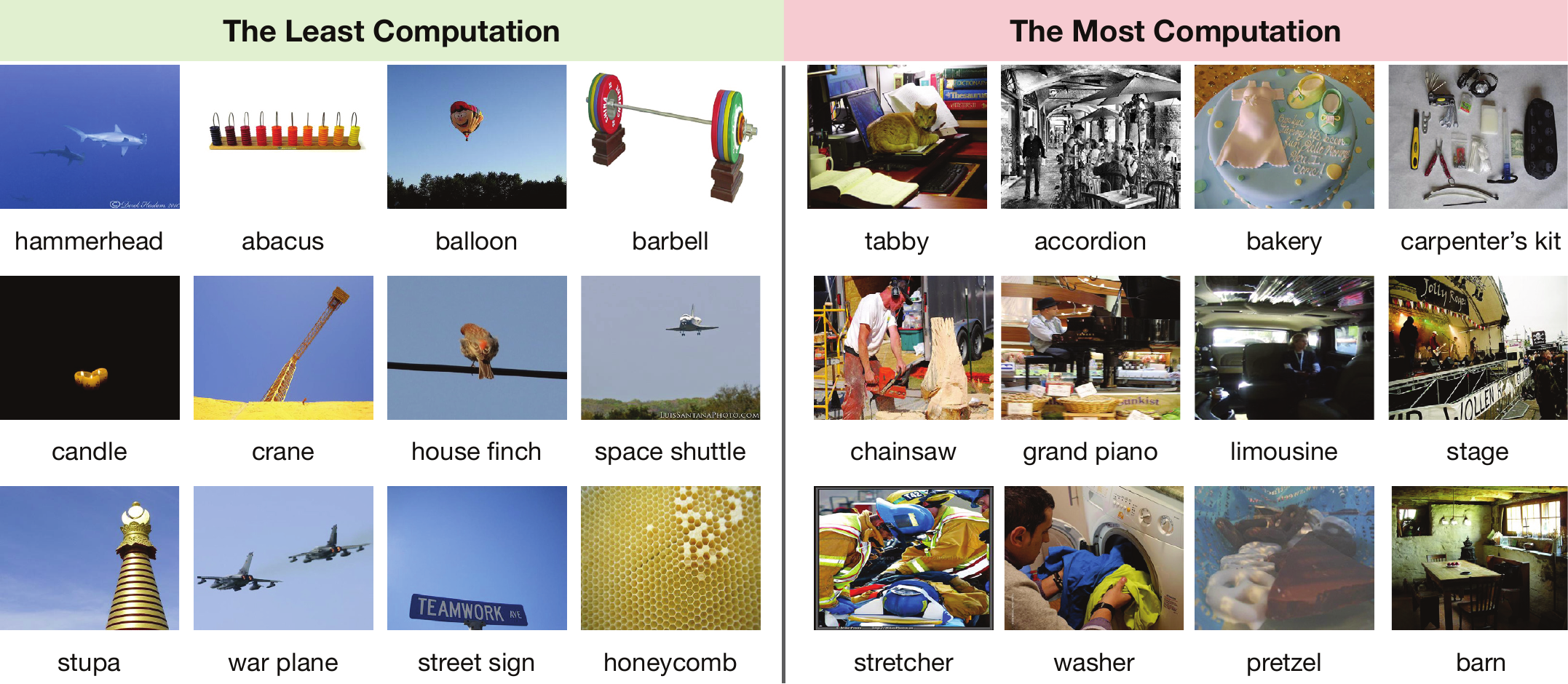}}
 \caption{\textbf{Qualitative results.} Images allocated with the least (\textbf{Left}) and the most (\textbf{Right}) computational resources by AdaViT are shown.}
\label{fig:supp_qual}
\end{figure*}

\section{Conclusion}

In this paper we presented AdaViT, an adaptive computation framework that learns which patches, self-attention heads and blocks to keep throughout the transformer backbone on a per-input basis for an improved efficiency for image recognition. To achieve this, a light-weight decision network is attached to each transformer block and optimized with the backbone jointly in an end-to-end manner. Extensive experiments demonstrated that our method obtains more than $2\times$ improvement on efficiency with only a small drop of accuracy compared with state-of-the-art vision transformers, and covers a wide range of efficiency/accuracy trade-offs. We further analyzed the learned usage policies quantitatively and qualitatively, providing more insights on the redundancy in vision transformers.

\appendix

\section{Qualitative Results}

We further provide more qualitative results in addition to those in the main text. Images that are allocated the least/most computational resources by our method are shown in Figure~\ref{fig:supp_qual}, demonstrating that our method learns to use less computation on easy object-centric images and more computation on hard complex images with cluttered background. Figure~\ref{fig:supp_patch} shows more visualization of the learned usage policies for patch selection, demonstrating the pattern that our method allocates less and less computation gradually throughout the backbone network, which indicates that more redundancy in computation resides in the later stages of the vision transformer backbone.

\begin{figure}[!t] \centering
    \resizebox{\linewidth}{!}{\includegraphics[width=\linewidth]{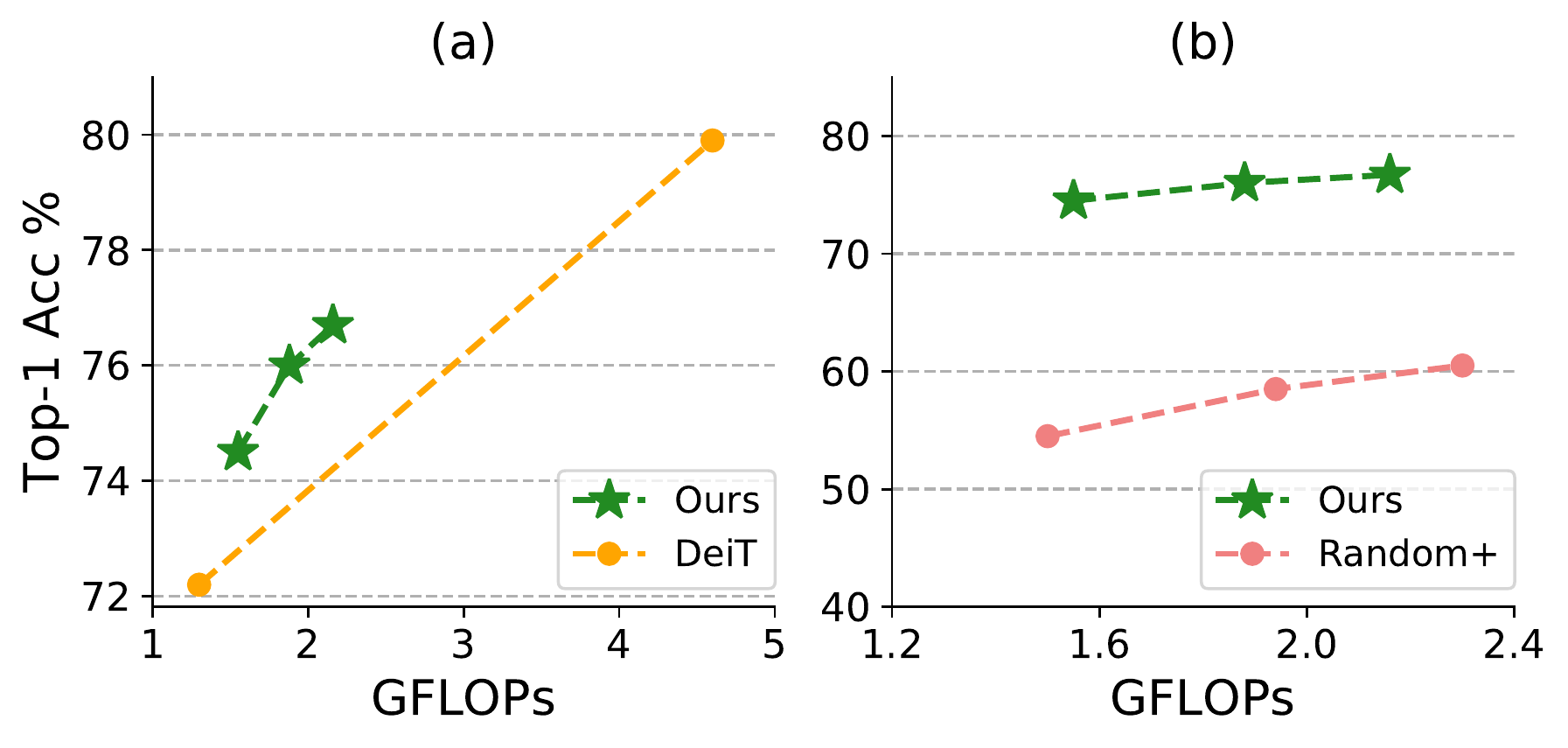}}
    \caption{\textbf{Compatibility to DeiT~\cite{touvron2021training_deit}.} We use DeiT-small as the backbone of AdaViT and show: \textbf{(a)} Efficiency/Accuracy tradeoffs of standard DeiT variants and our AdaViT. \textbf{(b)} Comparison between AdaViT and its \emph{Random+} baseline with similar computational cost.}
\label{fig:supp_deit}
\end{figure}

\section{Compatibility to Other Backbones}

\begin{figure*}[!th] \centering
    \resizebox{0.85\linewidth}{!}{\includegraphics[width=\linewidth]{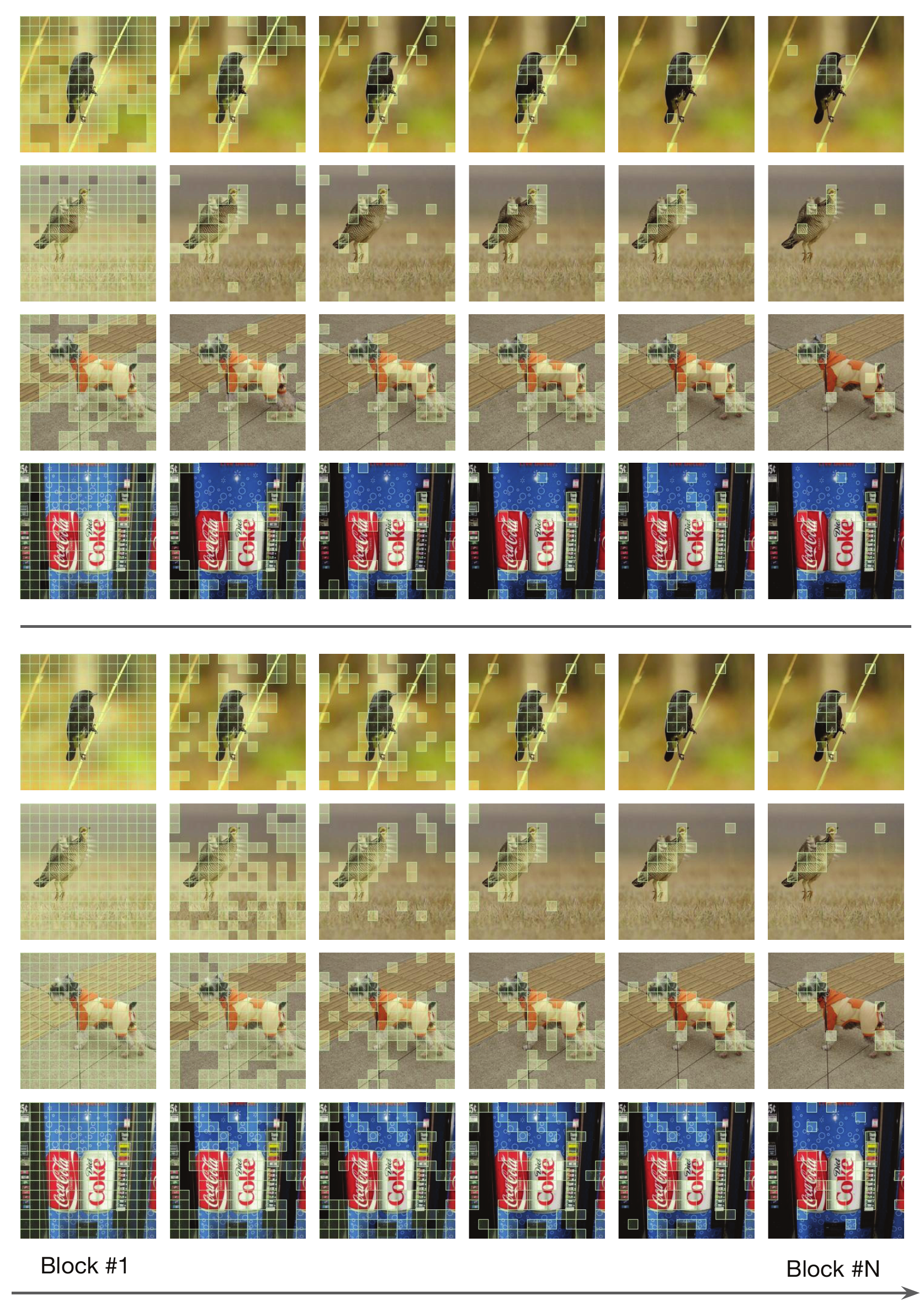}}
    \vspace{-2mm}
    \caption{\textbf{Visualization of selected patches at different blocks} with T2T-ViT~\cite{t2t} (\textbf{Above}) or DeiT~\cite{touvron2021training_deit} (\textbf{Below}) as the vision transformer backbone respectively. Green color denotes the patch is kept.}
\vspace{-5mm}
\label{fig:supp_patch}
\end{figure*}

Our method is by design model-agnostic and thus can be applied to different vision transformer backbones. To verify this, we use DeiT-small~\cite{touvron2021training_deit} as the backbone of AdaViT and show the results in Figure~\ref{fig:supp_deit}. AdaViT achieves better efficiency/accuracy tradeoff when compared with standard variants of DeiT, and consistently outperforms its \emph{Random+} baseline by large margins, as demonstrated in Figure~\ref{fig:supp_deit}(a) and~\ref{fig:supp_deit}(b) respectively. 

We further show the visualization of patch selection usage policies with DeiT-small as the backbone as well in Figure~\ref{fig:supp_patch}. A similar trend of keeping more computation at earlier layers and gradually allocating less computation throughout the network is also observed.

%%%%%%%%% REFERENCES
{\small
\bibliographystyle{ieee_fullname}
\bibliography{reference}
}

\end{document}